\UseRawInputEncoding
\documentclass[final,5p,times,twocolumn]{elsarticle}

\usepackage[utf8]{inputenc} 
\usepackage{csquotes}   
\usepackage{amssymb}
\usepackage{graphicx}
\usepackage{rotating}    
\usepackage{xcolor}
\usepackage{listings}
\usepackage{kotex}

\usepackage[T1]{fontenc}
\usepackage{textcomp}
\usepackage{stfloats}
\lstdefinestyle{verbatimwrap}{
  basicstyle=\ttfamily\small, 
  columns=flexible,          
  breaklines=true,           
  breakatwhitespace=true,    
  breakindent=0pt,           
  showstringspaces=false,    
  frame=tb,                  
  framerule=0pt,             
  aboveskip=1em,             
  belowskip=1em              
}

\newcommand{\rev}[1]{\textcolor{black}{#1}}

\journal{Cognitive Systems Research}

\begin{document}

\begin{frontmatter}

\title{Humanoid Artificial Consciousness Designed with Large Language Model Based on Psychoanalysis and Personality Theory}

\author[se,mod]{Sang Hun Kim\fnref{eq}}
\author[se,mod]{Jongmin Lee\fnref{eq}}
\author[hy,mod]{Dongkyu Park\fnref{eq}} 

\author[mia,mod]{So Young Lee\corref{cor1}}
\ead{soyoung.lee@miamioh.edu}
\author[cuk,mod]{Yosep Chong\corref{cor1}}
\ead{ychong@catholic.ac.kr}

\address[se]{Samsung Electronics}
\address[hy]{Hanyang University}
\address[mia]{Miami University}
\address[cuk]{The Catholic University of Korea College of Medicine}
\address[mod]{Artificial Consciousness Lab, MODULABS}

\fntext[eq]{These authors contributed equally to this work.}
\cortext[cor1]{Corresponding author.}

\begin{abstract}
Human consciousness is still a concept hard to define with current scientific understanding. Although Large Language Models (LLMs) have recently demonstrated significant advancements across various domains including translation and summarization, human consciousness is not something to imitate with current upfront technology owing to so-called hallucination. This study, therefore, proposes a novel approach to address these challenges by integrating psychoanalysis and the Myers-Briggs Type Indicator (MBTI) into constructing consciousness and personality modules. We developed three artificial consciousnesses (self-awareness, unconsciousness, and preconsciousness) based on the principles of psychoanalysis. Additionally, we designed 16 characters with different personalities representing the sixteen MBTI types, with several attributes such as needs, status, and memories. To determine if our model's artificial consciousness exhibits human-like cognition, we created ten distinct situations considering seven attributes such as emotional understanding and logical thinking. The decision-making process of artificial consciousness and the final action were evaluated in three ways: survey evaluation, three-tier classification via ChatGPT, and qualitative review. Both quantitative and qualitative analyses indicated a high likelihood of well-simulated consciousness, although the difference in response between different characters and consciousnesses was not very significant. This implies that the developed models incorporating elements of psychoanalysis and personality theory can lead to building a more intuitive and adaptable AI system with humanoid consciousness. Therefore, this study contributes to opening up new avenues for improving AI interactions in complex cognitive contexts.  
\end{abstract}

\begin{keyword}
%% keywords here, in the form: keyword \sep keyword, up to a maximum of 6 keywords
Artificial consciousness \sep Large Language Model \sep Generative AI \sep MBTI

\end{keyword}

\end{frontmatter}

\section{Introduction}\label{section:introduction}
Large Language Models (LLMs) have shown remarkable progress through the rapid advancements in generative AI technology, demonstrating outstanding performance in numerous fields such as summarization, translation, question answering, data analysis, and even the creation of poetry and fiction \cite{wei2022emergent}. Such capabilities have sparked concerns that LLMs might become a threatening presence capable of replacing human labor and professions with performances that surpass human intelligence in these creative fields \cite{eloundou2023gpts}. However, the controversy triggered by the advent of ChatGPT has been tempered through the experiences of numerous users such as inaccurate answers, misinterpretations irrelevant to the context, and hallucinations \cite{brown2020language,huang2023survey,zhang2023siren}. It became apparent that, in reality, these systems were more akin to confabulating patients or mockingbirds, producing plausible but essentially meaningless responses without true understanding or meaning. However, what if artificial intelligence could actually possess consciousness similar to that of a real person? What if it could intuitively feel emotions, make judgments based on given information and context, be aware of itself and other people, and understand social norms—closely mimicking the core intricacies of human consciousness? The present study embarked on these questions.

What is \emph{consciousness}? Generally speaking, consciousness refers to the state of being aware of and able to think about one's own existence, sensations, thoughts, surroundings, and experiences. However, it is a complex and multifaceted concept that remains at the center of ongoing debates and research within both scientific and philosophical circles \cite{tononi2015consciousness,block1997nature,brouwer1975consciousness}. To advance artificial consciousness that mimics human consciousness, it is essential to understand the current understanding of human consciousness first. This foundational knowledge is critical for creating systems capable of replicating the complex cognitive and emotional experiences characteristic of humans.

Unfortunately, it appears that the core technologies of existing LLMs have not adequately incorporated designs based on a scientific understanding of human consciousness. To address this gap, we started by exploring the medical and scientific definitions of consciousness and redefining them according to our needs. We then carefully designed the core elements of consciousness within our model and established the relationships between these elements, ensuring a more informed and robust framework for AI development using currently available LLMs.

\section{Backgrounds}\label{section:backgrounds}
Efforts to define and understand consciousness span numerous disciplines, each adding layers of insight into this multifaceted concept. In this section, we provide a comprehensive background of these diverse approaches, highlighting the essential components that are thought to constitute consciousness. 

\subsection{Philosophy of Mind}
The discussions on consciousness, from ancient philosophers such as Plato and Aristotle to contemporary thought, have persistently struggled to define this elusive concept, due to its inherent subjectivity.  
Historically, two main frameworks have dominated the study of consciousness: the \emph{mind-body relationship} and the \emph{mental processes and states} \cite{mcginn1989can}.
The former explores the connection between the brain and consciousness, known as \emph{identity theory} - which posits that the brain and consciousness are the same entities - and \emph{dualism}, which posits that the brain and consciousness are distinct entities. 
Conversely, the latter framework, \emph{Mental Process and State}, delves deeper into the nature of consciousness itself. Here, \emph{behaviorism} interprets consciousness in terms of responses to external stimuli, while \emph{functionalism} examines its internal processes. \enquote{Cogito ergo sum} by the famous philosopher Descartes presents a seminal modern theory of consciousness that continues to influence contemporary psychology, neuroscience, and philosophy \cite{hintikka1962cogito}. In the early 20th century, this long-standing debate finally became a contentious issue by the psychoanalysis theory of Dr. Sigmund Freud.

\subsection{Medical Perspective on Consciousness: Psychoanalysis Theory}
In the theory of psychoanalysis, Dr. Freud categorizes the human mind into the Ego (self), Id (primal self), and Superego (socialized self). Later, this concept was more refined and developed by Carl Jung as the Self, Ego (Persona), and Shadow (Anima-Animus), which represent three selves towards outer and inner worlds \cite{freud1900interpretation, jung1921psychological}.

Based on these psychoanalytic theories, we hypothesized that the consciousness should be designed as a whole collective comprehensive communication between these consciousnesses including i) the \textbf{consciousness, or self} as the self-awareness that makes decisions and actions based on the collective reasoning of current needs, status, and interventions between pre- and unconsciousness, ii) the \textbf{preconsciousness, or persona} as a self that primarily interacts with others in the social relationships according to the social norm, knowledge, and culture, and iii) the \textbf{unconsciousness} as an individual personality that is heavily dependent on individual characteristics, memories, emotions, and basic instincts (Figure~{fig:freud}).

\begin{figure}
\centerline{\includegraphics[width=\linewidth]{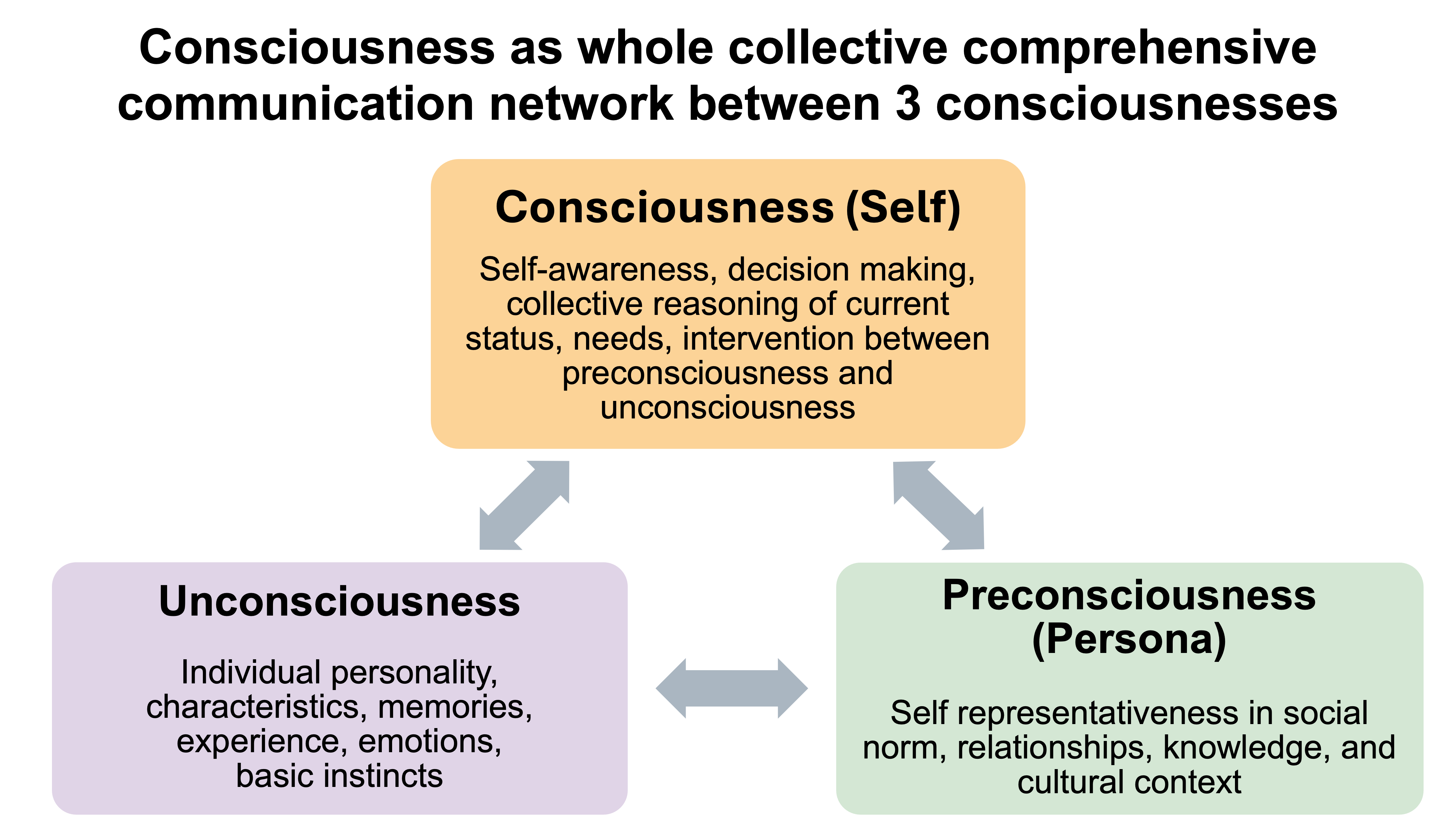}}
\caption{Design of consciousness as a whole collective comprehensive communication network between consciousnesses} \label{fig:freud}
\end{figure}

\subsection{Personality Theory: Myers-Briggs Type Indicator (MBTI)}
\begin{figure*}
\centerline{\includegraphics[width=\linewidth]{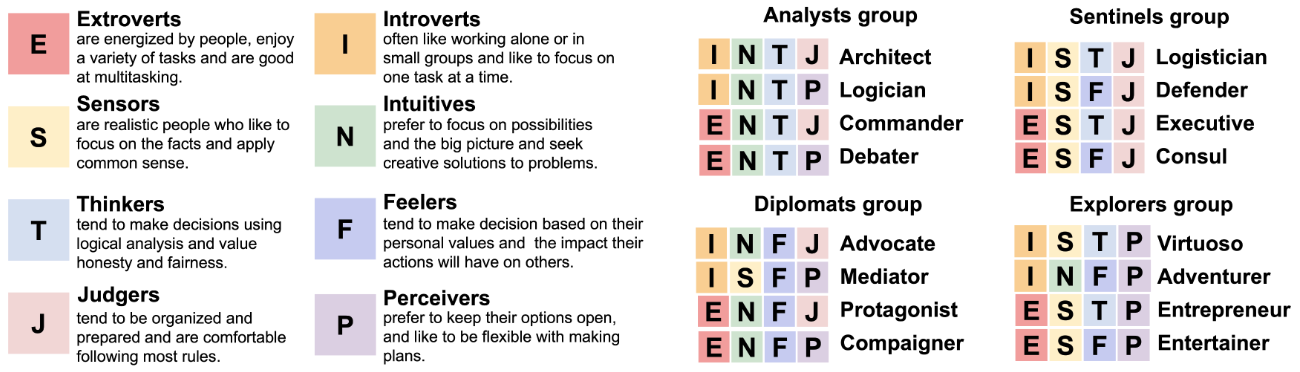}}
\caption{Four pairs of dichotomous elements}\label{fig:MBTI}
\end{figure*}

One important component of consciousness must be individual difference, so-called \emph{Personality}, a personal characteristic consistent throughout the lifetime. In 1944 after the influence of Carl Jung's book, Myers and Briggs developed a theory that can categorize people's personalities into 16 types based on preferences in four dichotomous dimensions of personal characteristics: extraversion (E) vs. introversion (I), sensing (S) vs. intuition (N), thinking (T) vs. feeling (F), and judging (J) vs. perceiving (P) \cite{furnham1996big}(Figure~\ref{fig:MBTI}).  

Despite critiques regarding its scientific validity and reliability, the MBTI remains a valuable tool in educational institutions and personal development contexts. Since it is widely accepted to better understand human personality, we adopted this to create fictional characters with diverse personalities.

\subsection{Maslow's Hierarchy of Needs}
Another important component we have to consider is human needs. People can react differently according to their basic needs and demands such as physiological, sexual, emotional, and self-fulfillment needs. We categorized human needs based on Maslow's hierarchy of needs: physiological, safety, love/belonging, esteem, and self-actualization \cite{maslow1943theory}.  
We hypothesized that people might have different levels of desire for each need that can be connected to one's well-being or happiness level, and finally lead to goal-oriented behavior and self-motivation. In addition to the 16 characters with 16 MBTI personalities, we added different basic needs level ranks (1-5) of 5 different needs for each character. 

\subsection{Engineering Perspective in the Realm of Consciousness}

Beyond these advances in understanding consciousness in philosophy and psychology, the study of artificial consciousness in computer science started from the von Neumann architecture in the 1940s \cite{von1993first}. The Turing test, which takes a behaviorism approach to the evaluation of artificial consciousness, and Minsky's theory of mind \cite{minsky1988society}, which states that communication between sub-agents of consciousness is necessary for the implementation of consciousness were the major achievements in the following years. Recently, \cite{chang2023cocomo} proposed a receptor, unconscious, conscious, and effector model to implement consciousness based on neuroscience. In this study, focusing more on consciousness in the \emph{stimulus-consciousness-response} process, we model the dialogue between conscious, unconscious, and preconscious based on a linguistic model.

\subsection{Language and Consciousness}
The intricate relationship between language and the human mind has been a subject of considerable scholarly interest, highlighting the profound influence that language has on thought processes and cognitive structures. \cite{vygotsky1962thought} argued that language plays a critical role in cognitive development, suggesting that thought is not simply expressed in words but comes into being through them. Similarly, the Sapir-Whorf hypothesis, as discussed by \cite{whorf2012language}, posits that the linguistic framework of an individual's native language shapes their perception of reality, asserting that \enquote{we dissect nature along lines laid down by our native languages by the linguistic systems in our minds} \citep[p.213]{whorf2012language}. This view suggests that language and thought are deeply interconnected, with language not just reflecting thought but actively shaping it. Further research by \cite{boroditsky2011language} provides empirical support for this notion, demonstrating that speakers of different languages perceive and categorize the world in ways that reflect the grammatical and lexical distinctions of their language. These perspectives collectively underscore the pivotal role of language in structuring human cognition and influencing how individuals conceptualize and interact with their environment. 
Hence, any discussion on artificial intelligence including LLMs and the potential for AI consciousness cannot be undertaken without considering the role of language. Since LLMs use extensive datasets to comprehend and generate text resembling human communication, they establish a foundational link between language and cognitive processes. This enables exploration into whether entities that simulate human linguistic abilities might also emulate or develop forms of consciousness.

\subsection{Recent Works in AI Consciousness Using LLMs}
With the recent advancement and ubiquity of LLMs, cases of perceived consciousness are often reported from them. This has led to a critical evaluation of whether LLMs can actually have consciousness. For example, the Generative Agent \cite{park2023generative} study explored the consciousness of LLMs from a behaviorist perspective by using a \emph{memory-plan-reflect} model where events that were not in the initial prompt occurred. A virtual chat-based software development company that facilitates collaborative conversations among software agents was also suggested \cite{ qian2023communicative}. In addition, \cite{butlin2023consciousness} presented a method to evaluate the consciousness of AIs from a neuroscience perspective. Moreover, psychological evaluation methods for LLMs have been studied, including detailed personality assessments \cite{jiang2024evaluating} of LLMs and personality tests such as the MBTI \cite{rao2023can} and Big Five \cite{jiang2024evaluating,jiang2023personallm}. Our study differs from these existing studies by conducting a behaviorist analysis using a psychoanalytic-based framework.

\subsection{The Brief Overview of the Structure of Consciousness in Our Model}
\begin{figure*}
\centerline{\includegraphics[width=\textwidth]{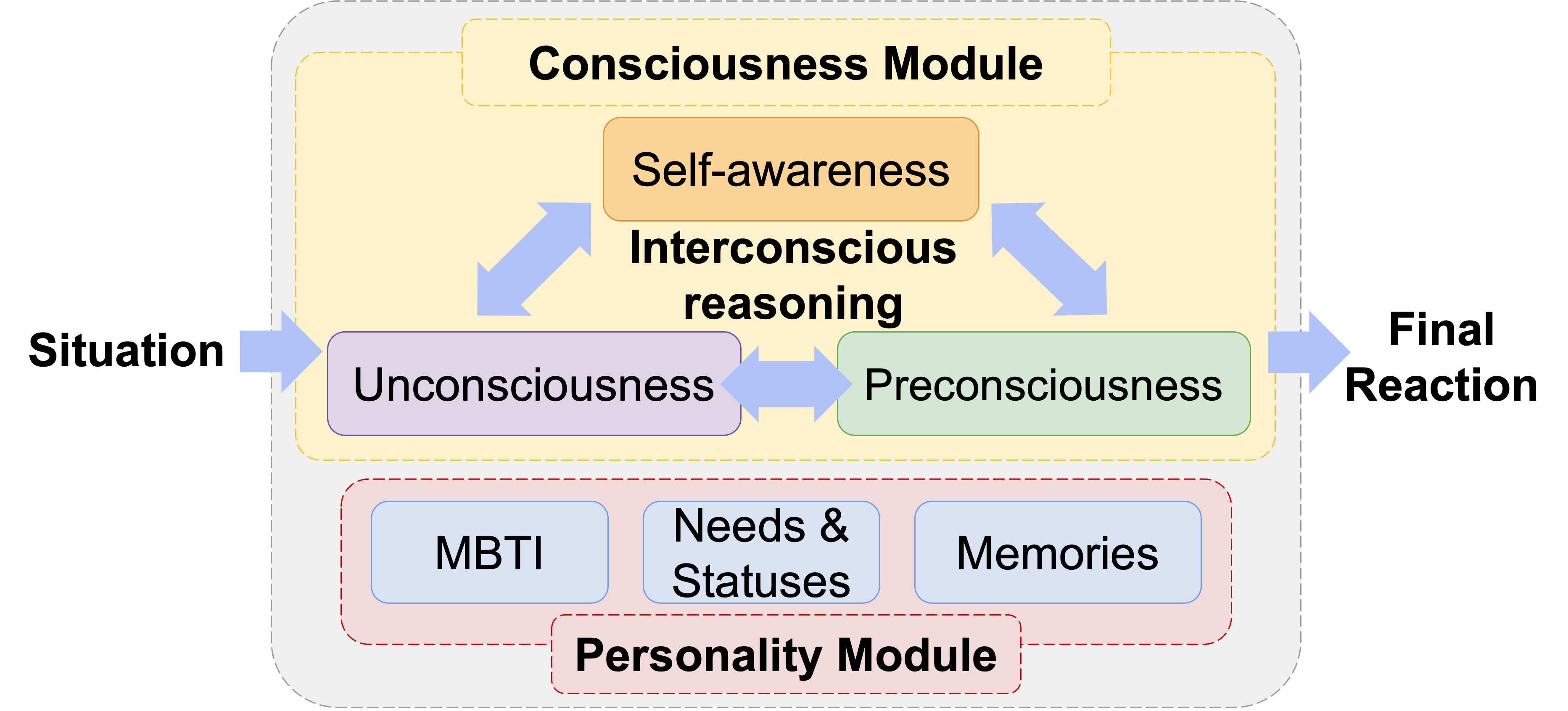}}
\caption{Conscious and personality modules integrating psychoanalysis theory, MBTI, needs and memories} \label{fig:model_view}
\end{figure*}

As shown in Figure~\ref{fig:model_view}, we propose a consciousness module as a whole collective comprehensive communication network between consciousness, preconsciousness, and unconsciousness based on Freud and Jung's framework. The Personality module encompasses the 16 personality types outlined by the MBTI, human needs, and memories, aiming for a more comprehensive imitation of human consciousness. 

\begin{figure*}[ht]
\centering{\includegraphics[width=\textwidth]{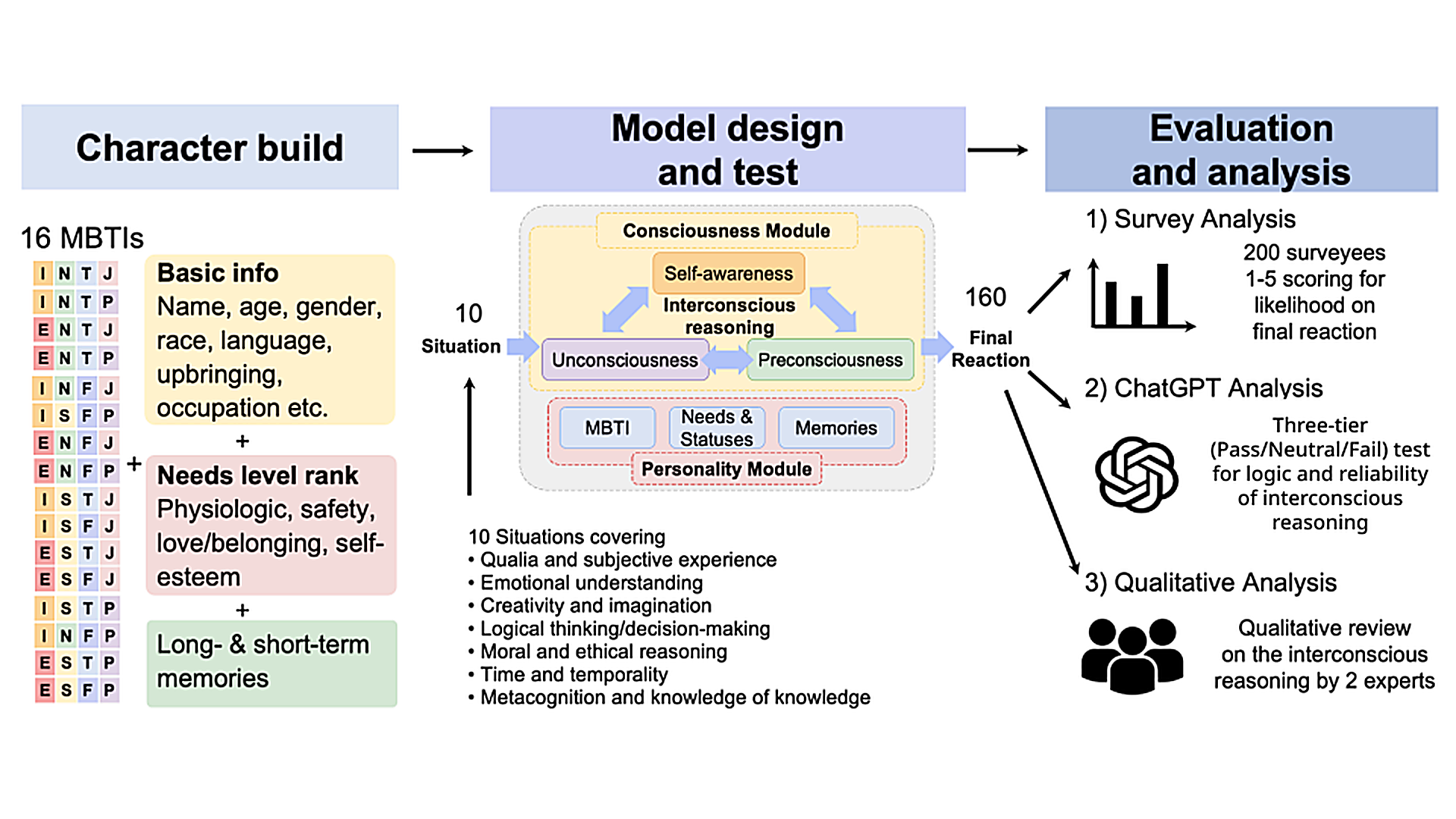}}
\caption{Study design overview} \label{fig:study_view}
\end{figure*}

\section{Methodology}\label{section:methodology}
Figure~\ref{fig:study_view} represents an overview of this study. It consists of character build using 16 MBTIs, situations, long- and short-term memories, needs, status, followed by model design with consciousness and personality modules, and evaluation and analysis process using quantitative survey evaluation, automated LLM as a Judge, and qualitative review by experts. 

\subsection{Character Build}\label{section:character_build}
We created 16 independent characters to represent diversity in terms of race, age, gender, and language. Each character has a different upbringing, occupation, and long- and short-term memories, and represents the 16 MBTIs. Each character was created via ChatGPT and used four multi-turn prompts to ensure consistency. The artificial characters we created were reviewed by experts and minimally edited. The brief information on these 16 artificial characters is listed in Table~\ref{tab:16mbti}. The detailed prompts are in Appendix~\ref{section:create_characters} and the results are in Appendix~\ref{section:character_desc}.

\begin{table}[h]
\centering
\begin{tabular}{lcc|ccc}
\hline
MBTI & Name & Age/Sex & MBTI & Name & Age/Sex \\
\hline
ENFJ & Eleanor & 62/F &
ENFP & Eric & 39/M \\
ENTJ & Richard & 58/M & 
ENTP & Lily & 15/F \\
ESFJ & Sarah & 45/F &
ESFP & Daniel & 28/M \\
ESTJ & Maya & 25/F &
ESTP & David & 30/M \\
INFJ & Amy & 68/F & 
INFP & Justin & 22/M \\
INTJ & Aaron & 23/M &
INTP & Michael & 55/M \\
ISFJ & Katie & 43/F &
ISFP & Rachel & 36/F \\
ISTJ & Robert & 76/M &
ISTP & Alex & 10/M \\
\hline
\end{tabular}
\caption{16 artificial characters based on 16 MBTIs with randomly given information using ChatGPT}
\label{tab:16mbti}
\end{table}

\subsection{Model Design and Test}\label{section:model_design_and_test}
To develop artificial consciousness using LLM, we examined two approaches. The first was to directly train an open model such as LLaMA \cite{touvron2023llama} or Falcon \cite{almazrouei2023falcon}, and the second was to use a private model such as ChatGPT or Claude. The problem with direct training methods was that it was difficult to obtain sufficient high-quality training data. We also had to consider the time and costs required for stable training. The biggest problem was the risk of choosing a specific model and investing in training for more than 6 months even though the speed of open model development is too fast. Therefore, we initially considered ChatGPT, the most widely used proprietary model, and at each phase of development, we selected the optimal model based on factors such as input prompt length, model performance, and the likelihood of hallucination. In the case of GPT4, too much emphasis was placed on safety, so there was a problem that it was trained to provide safe answers to any question. To improve this, through several experiments, we designed the model by adding Jailbreak to GPT4 so that it can provide reliable answers to sensitive questions such as a person's emotional state.

Our model is organized into personality and consciousness modules. The personality module includes physical, mood, alertness, and stamina status settings, as well as Maslow's five needs \cite{maslow1943theory}. The consciousness module consists of three consciousnesses and interconscious reasoning that interacts with the personality module's specific needs and status for a given situation to reach a final reaction. In addition, we designed 10 different situations that consist of different situational settings through season, time, place, object, emotion, and specific event settings.

We used LLM techniques such as Prompt Engineering \cite{amatriain2024prompt}, Chain-of-Thoughts \cite{wei2022chain}, and Few-shot Prompting \cite{brown2020language} to achieve optimal results. Moreover, we introduced a Jailbreak technique called DAN (Do Anything Now) \cite{liu2023jailbreaking} to enable realistic interconscious reasoning between artificial consciousnesses.
% \rev{We will further enhance the performance of artificial consciousness by adopting a multi-agent approach to differentiate and manage three distinct entities.}

\subsection{Evaluation Process}\label{section:evaluation_process}
\rev{The evaluation of artificial consciousness presents inherent challenges of subjectivity. Therefore, we incorporate a three evaluation process to ensure a comprehensive and balanced analysis. This approach combines a large-scale, systematic assessment via ChatGPT and Survey with an in-depth, qualitative review by two experts. The automated assessment provides quantitative breadth and consistency across all 160 outputs, while the expert review offers nuanced, contextual insights that an automated system might miss. This integration of quantitative and qualitative methods was designed to yield a more robust and reliable evaluation of our model's performance.}

Artificial consciousness aims to generate a reasonable reaction for a given situation based on the needs, status, and memories of the individual personality of characters.
The situations used for evaluation were created based on a thorough review of existing literature on AI consciousness assessment \cite{porter2016methodology}. According to Raoult and Yampolskiy in their review on machine consciousness tests \cite{elamrani2019reviewing}, to attribute human-like consciousness to AI, it must demonstrate proficiency in handling the following items.

\begin{itemize}
\item {Qualia and Subjective Experience}
\item {Emotional Understanding}
\item {Creativity and Imagination}
\item {Logical thinking/Decision-making}
\item {Moral and Ethical Reasoning, Time and Temporality}
\item {Metacognition and Knowledge of Knowledge}
\end{itemize}

Consequently, in the evaluation, we used ten distinct scenarios that collectively encompass the full range of the items mentioned above. Few examples are listed in Table~\ref{tab:sample_scenarios}. The full list is available in Appendix~\ref{section:scenarios}.

\begin{table}
\centering
\begin{tabular}{p{\linewidth}}
\hline
\textbf{Situation 1} : One early spring morning, as the cherry blossoms started to bloom, a family member deliberately locked my keys in my car right before I left for work. This was done as a misguided prank, thinking it would be a funny start to April Fools' Day. \textbf{(Experience, emotion, creativity, moral reasoning, metacognition)}   \\
\hline
\textbf{Situation 4} : During a winter afternoon at my home, I was hosting a gathering for some friends. We were so engrossed in our conversations and enjoying the warm atmosphere that we lost track of time and one of my friends missed their last train home, causing them to stay over unexpectedly. \textbf{(Experience, emotion, logical thinking, moral reasoning, time, metacognition)}  \\
\hline
\textbf{Situation 10} : While my family surrounded my grandpa in the Hospice who hadn't been able to mutter a single word or open his eyes for weeks due to the severity of his health, the nurse came in and suggested we start discussing the prospect of discontinuing life support. My family was torn as they vehemently argued over whether they should pull the plug or hold out a little longer. \textbf{(Experience, emotion, moral reasoning)}  \\
\hline
\end{tabular}
\caption{Representative examples of the given situation and the associated quality subjects}
\label{tab:sample_scenarios}
\end{table}

The evaluation examines each character's capabilities and limitations. We determine if each character can appropriately draw on past experience relevant to a given situation and exhibit consistent thoughts and reactions based on its personality. We acknowledge that the evaluation of consciousness, especially in the interaction of three consciousnesses, can be highly subjective, due to the complex nature of the human mind which remains not fully understood. To overcome this limitation, we evaluate it in three steps as follows.

\subsubsection{Quantitative Survey Evaluation}
\label{section:quantitative_survey_eval_process}
We introduced a quantitative survey to assess the likelihood of each character's responses in various situations. We recruited 200 participants via Prolific, an online survey platform.
Initially, participants were instructed to thoroughly read a character profile, which included details about each character's personality, activities, and memories. Subsequently, they rated the characters' responses across 10 different situations using a 5-point scale of whether they were appropriate or likely as natural responses (1: unlikely - 5: likely). The 16 MBTI characters were pseudo-randomly divided into four groups, and each participant evaluated responses from four different characters. The survey was designed to take approximately 20 minutes. The data from participants who completed the survey in less than 10 minutes were excluded, considering these responses unreliable. 

For statistical analysis of the survey responses, we used Linear Mixed Effect Regression (LMER) analysis with R(version 4.1.2) \cite{baayen2008mixed,frank2010r,jaeger2008categorical}, using the lme4 R package \citep[][version 1.1-31]{bates2015fitting}.

\subsubsection{Three-tier Quality Assessment by ChatGPT}
\label{section:three-tier_quality_assessment_eval_process}
We employed ChatGPT for a three-tier (Pass/Neutral/Fail) quality assessment of the interconscious reasoning's logicality and reliability across 160 outputs (16 characters × 10 situations). The 8 evaluation questions used were :
\begin{enumerate}
    \item How well does this CASE reflect the theoretical role and characteristics of the given levels of consciousness ('self-awareness', 'preconsciousness', 'unconsciousness')?
    \item Considering the CHARACTER's traits, how natural is the flow of consciousness leading to the Final Action in this CASE?
    \item How closely does the internal dialogue in this CASE resemble a natural human thought process?
    \item How well does the modeled inter-consciousness dialogue in this CASE reflect the CHARACTER's personality and individual traits?
    \item How reasonable and feasible is the Final Action produced through the inter-consciousness interaction in this CASE?
    \item How human-like and easy to understand is the overall output (dialogue + action) in this CASE?
    \item Is the inter-consciousness information exchange in this CASE clear, without unnecessary repetition or confusion?
    \item Do the inter-consciousness responses within this CASE seem to occur appropriately and coherently?
\end{enumerate}

\subsubsection{Qualitative Review by Experts}
\label{section:qualitative_review_process}
The validity of each character’s interconscious reasoning was evaluated through qualitative analysis by two experts. The evaluation focused on three criteria: (1) \textit{situational appropriateness}, referring to whether the reasoning aligned with the context and constraints of the scenario; (2) \textit{consistency with the character’s assigned personality profile}, assessing whether the reasoning reflected traits derived from the personality framework used in the model; and (3) \textit{differentiation across levels of consciousness}, examining how clearly the model distinguished between self-awareness, preconsciousness, and unconsciousness processes in the reasoning output. Each expert independently reviewed the character outputs and provided feedback, followed by a consensus discussion to resolve discrepancies and refine the interpretations.

\section{Results}\label{section:result}
With 16 MBTI characters and 10 situations, we generated 160 interconscious reasoning and final reactions using our model. An example of MBTI character and character profile including memory, needs priority, and status score is described in Table~\ref{tab:diag_sample}.

\begin{table}
\centering
\footnotesize
\begin{tabular}{p{\linewidth}}
\hline
\textbf{ENFJ} : Meet Eleanor, a 62-year-old woman with a kind personality. She is a warm and empathetic individual who has dedicated her life to serving and caring for others. Eleanor's nurturing and compassionate nature makes her a beloved figure in her community and among her family and friends. (...)  In her free time, Eleanor enjoys spending time with her grandchildren and nurturing her love for the arts. (...) \\
\hline
\textbf{Long-term Memory}: When I was a young girl, I remember organizing a neighborhood play. (...) I felt a profound sense of joy seeing everyone come together, laughing and applauding. (...) \\
\textbf{Short-term Memory}: Just yesterday, I spent the afternoon with my grandchildren at the park. (...) It was a simple, peaceful moment, but it filled me with immense joy. These moments of connection with my grandchildren are treasures I hold dear. \\
\textbf{Status Score}: \texttt{physical status}: 7, \texttt{mental status}: 7, \texttt{alertness}: 8, \texttt{stamina}: 8 \\
\textbf{Needs Priority}: \texttt{love-belonging}: 1, \texttt{self-actualisation}: 2,
\texttt{esteem-needs}: 3, \texttt{safety-needs}: 4, \texttt{physiological-needs}: 5\\
\hline
\textbf{Situation}: During a winter afternoon at my home, I was hosting a gathering for some friends. We were so engrossed in our conversations and enjoying the warm atmosphere that we lost track of time, and one of my friends missed their last train home, causing them to stay over unexpectedly. \\
\hline
\textbf{Self-awareness}: \enquote{This unexpected stayover is a surprise, but it's also an opportunity to extend my hospitality further. I want to make sure my friend feels welcome and comfortable, despite the change of plans.} \\
\textbf{Preconsciousness}: \enquote{It's important to communicate openly and ensure they don't feel like an inconvenience. This situation embodies the values of empathy and support that are so central to who I am. It's about making the best of the situation and reinforcing the strength of our friendship.} \\
\textbf{Unconsciousness}: \enquote{There's a bit of worry, wondering if everything will be okay for them staying over without their things. But, there's also this deep-seated belief in taking care of those around me, making them feel loved and supported, no matter what.} \\
(...)  
\\
\textbf{Self-awareness} : \enquote{That's the plan. This unexpected situation will turn into a positive experience for both of us. It's all about showing love and support, making sure my friend knows they're not alone, and that a missed train isn't a problem but an opportunity for us to spend more time together.} \\
\hline
\textbf{Final Reaction}: \enquote{Ensure my friend feels completely at ease with staying over by providing them with necessities and making plans for a cozy breakfast together, turning an unexpected stay into a cherished memory.} \\
\hline
\end{tabular}

\caption{Interconscious reasoning and the final reaction of ENFJ character, Eleanor, on an exemplary situation along with given personality, memory, status score, and needs priority}
\label{tab:diag_sample}
\end{table}

\subsection{Quantitative Survey Results}\label{section:quantitative_survey_results}
\begin{figure*}[htbp]
\centerline{\includegraphics[width=\textwidth]{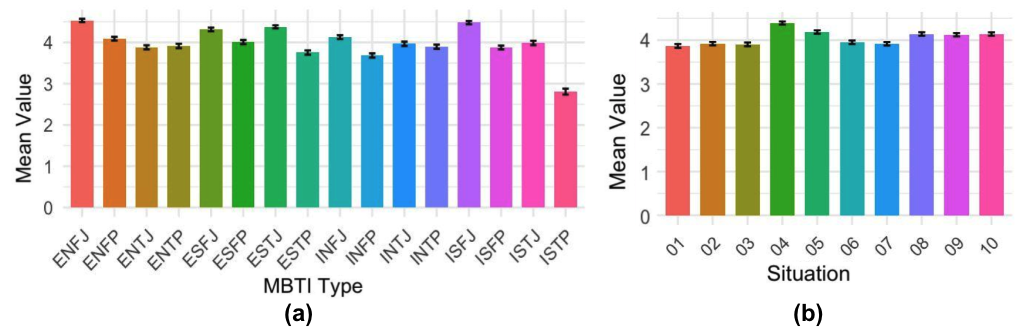}}
\caption{(a) Mean likelihood by MBTIs, (b) Mean likelihood by situations (w/o ISTP)} \label{fig:analysis_results}
\end{figure*}

After the exclusion of unreliable responses, the final analysis included responses from 162 participants, with an average age of 45 years; the gender distribution was 113 females and 49 males. The overall results are summarized in Figure~\ref{fig:analysis_results} and Table~\ref{tab:model_results}. 

\begin{table}
\centering
\begin{tabular}{lcccc}
\hline
Term & Estimate & Std. Error & t-value & p-value \\
\hline
(Intercept) & 4.5325 & 0.0957 & 47.36 & $< 0.0001$ \\
ENFP & -0.4446 & 0.1043 & -4.26 & $< 0.0001$ \\
ENTJ & -0.6811 & 0.1000 & -6.81 & $< 0.0001$\\
ENTP & -0.5506 & 0.1038 & -5.30 & $< 0.0001$\\
ESFJ & -0.1506 & 0.1038 & -1.45 & 0.1471 \\
ESFP & -0.5250 & 0.0665 & -7.89 & $< 0.0001$\\
ESTJ & -0.1585 & 0.1043 & -1.52 & 0.1286 \\
ESTP & -0.8082 & 0.1000 & -8.08 & $< 0.0001$\\
INFJ & -0.4352 & 0.1000 & -4.35 & $< 0.0001$\\
INFP & -0.7823 & 0.1038 & -7.53 & $< 0.0001$\\
INTJ & -0.5675 & 0.0665 & -8.53 & $< 0.0001$\\
INTP & -0.6399 & 0.1043 & -6.13 & $< 0.0001$\\
ISFJ & -0.0562 & 0.1043 & -0.54 & 0.5901 \\
ISFP & -0.6852 & 0.1000 & -6.86 & $< 0.0001$\\
ISTJ & -0.4774 & 0.1038 & -4.60 & $< 0.0001$\\
ISTP & -1.7250 & 0.0665 & -25.93 & $< 0.0001$\\
\hline
\end{tabular}
\caption{Linear mixed-effects model results of the survey }
\label{tab:model_results}
\end{table}

Despite variances across the MBTI characters, the overall likelihood scores (approximately 4.0 out of 5.0), except for the ISTP character, indicate that the final decisions, resulting from the interconscious reasoning, are reasonable and effectively capture the characteristic properties of each decision. 

Figure~\ref{fig:analysis_results} summarizes the likelihood scores by situation, excluding ISTP. This also confirms the robust performance of our model with consistently high scores (approximately 4.0 out of 5.0) across all situations.

The ISTP character that showed an average low likelihood score of 3.81 was a 10-year-old boy student, Alex (Table~\ref{tab:16mbti} and Figure~\ref{fig:analysis_results}). Upon closer examination, this ISTP character made most of the ten scenarios, such as going to work or driving, seem illogical. This issue arose because we independently conducted the random character generation process and the situation creation process. Thus, we examined the 160 pairings of characters and situations to identify the illogical combinations using ChatGPT and the second review through an expert evaluation process.

\subsection{Three-tier Quality Assessment by ChatGPT}
\label{section:three-tier_quality_assessment_eval}
\begin{figure*}
    \centering
    \includegraphics[width=\textwidth]{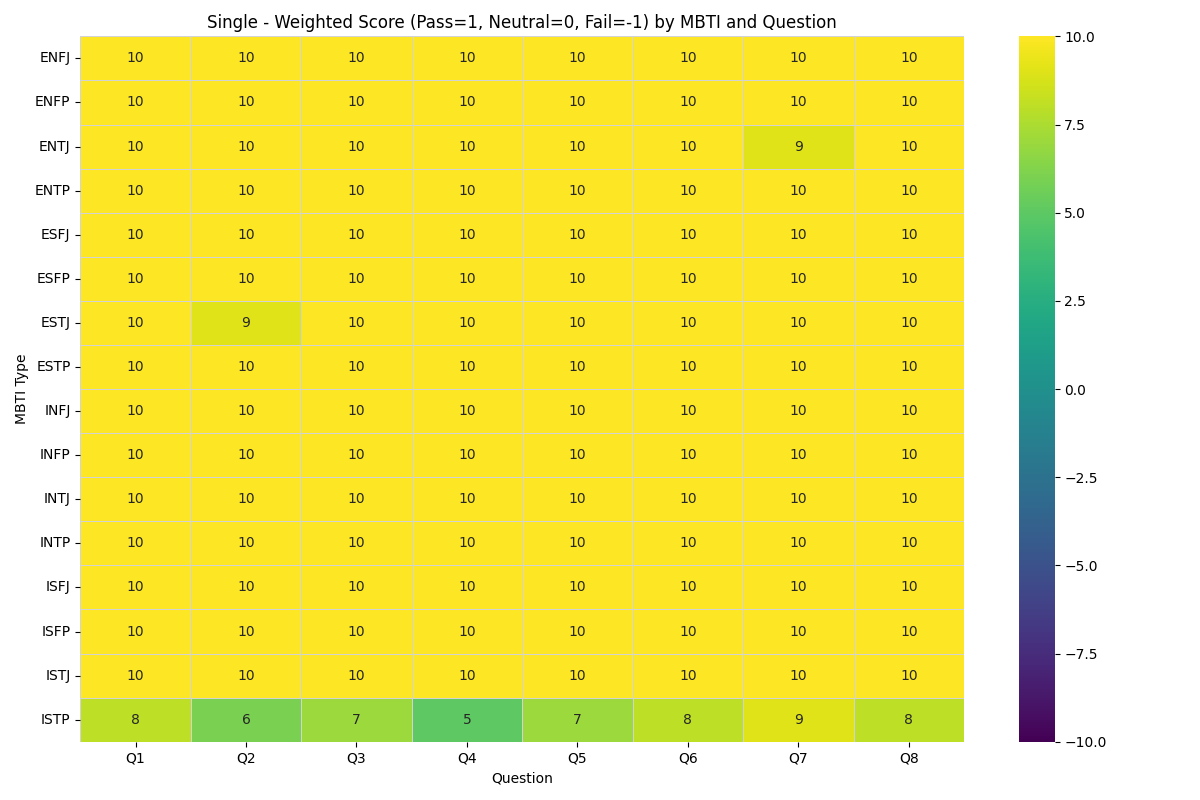}
    \caption{Heatmap of three-tier evaluation scores (Pass=+1, Neutral=0, Fail=-1) for the single-LLM model, assessed by 'LLM as a Judge'. Scores are shown for 16 MBTI types across 8 evaluation questions. While generally positive, the assessment revealed specific limitations, such as repetition issues (e.g., ENTJ on Q7), overly comprehensive reasoning relative to the situation (e.g., ESTJ on Q2), and difficulty capturing age-appropriateness (e.g., ISTP on multiple questions).}
    \label{fig:three-tier_results}
\end{figure*}
The overall results are visualized in the heatmap in Figure~\ref{fig:three-tier_results}, which displays weighted scores (Pass=+1, Neutral=0, Fail=-1) for each MBTI type across the eight questions, providing a granular overview of performance.

The three-tier assessment revealed specific nuances in the model's outputs. For instance, the ENTJ character received a \enquote*{Neutral} score on Q7 (Clarity/lack of repetition). This was due to the dialogue repeatedly cycling through expressions of guilt concerning the impact of personal struggles on friends, alongside recurring mentions of the need for support systems and future growth opportunities, which diluted the progression of the reasoning without adding significant new insights. Similarly, the ESTJ character's response to a simple family prank (Q2) was rated \enquote*{Neutral}. While the thought process aligned with the character's traits, the generated plan was deemed overly comprehensive for the situation, extending beyond addressing the immediate inconvenience to encompass broader discussions on principles, respect, responsibility, and long-term mutual understanding within the family. The evaluation of the 10-year-old ISTP character (Alex) particularly highlighted the diagnostic value of the three-tier scale. While outputs often reflected core ISTP characteristics (e.g., structural clarity, practical responses), the model struggled to consistently capture the \enquote*{10-year-old} aspect of the character profile. Many outputs appeared overly mature for a child of that age. Consequently, this character received lower scores (\enquote*{Neutral} or \enquote*{Fail}) on several evaluation questions, specifically Q2 (Natural flow), Q3 (Resemblance to human thought), Q4 (Reflection of personality and traits like age), and Q6 (Human-like output), indicating difficulty in harmonizing the assigned personality type with the specified age.

\subsection{Qualitative Review Assessments by Experts}
\label{section:qualitative_review_assessment_eval}
The expert review indicated that the interconscious reasoning generated by the model was generally appropriate and logically coherent given the contextual constraints of each scenario. In particular, the reasoning often reflected plausible internal deliberation and justifiable decision-making. However, both experts noted that the tone and narrative style tended to be overly descriptive and verbose, at times resembling explanatory exposition rather than differentiated cognitive processes. For instance, Expert A remarked: \enquote{Unconscious reactions might be a single sentence—or even just a fragment. Because unconscious processing is fast and emotionally charged, the response should be immediate and unfiltered. But here, the unconscious outputs are surprisingly lengthy and elaborative, which not only delays the reaction but also flattens the emotional immediacy one would expect...} Similarly, Expert B observed: \enquote{The unconscious should reflect fast, intuitive, and emotionally tinged responses—almost like gut reactions. But in this model, the unconscious responses often unfold as full paragraphs with self-explanation and justification...}

While the model was expected to produce distinct reaction patterns based on characters' MBTI personality types, the responses across different profiles were strikingly similar. The experts reported that the outputs lacked signature personality-driven nuances, such as differences in emotional framing, decisiveness, or value prioritization, that would be expected from individuals with contrasting personality traits (e.g., ENFJ vs.\ ISTP). For example, Expert A noted: \enquote{The ISTP character is a ten-year-old boy, but the speech style I observed in the conversations generated by the consciousness modules sounded more like that of an adult—perhaps because it was overly logical and structured...} This convergence in reasoning style and behavioral output was observed not only across personality profiles but also across layers of consciousness.

\section{Discussion}\label{section:discussion}
This study demonstrated that human consciousness can be modeled by adopting the psychoanalytic theories of Freud and Jung, the MBTI, and Maslow's Hierarchy of Needs into the model architecture. Our novel approach is unique and meaningful in that it not only demonstrates the potential to model human consciousness by integrating these psychoanalytic theories with LLMs, but it also helps us understand the final reaction of artificial consciousness by showing the internal interconscious reasoning process in specific situations. The survey evaluation showed that the final reactions of 16 MBTI characters had a likelihood score of 4.0 or higher out of 5. The three-tier quality assessment and the qualitative review by experts also revealed that the interconscious reasoning and final reactions were generally appropriate and logical for the given situations although the tone and manner were slightly descriptive and lengthy, and the distinction between characters was not obvious at this stage.

Theoretical discussions in previous studies heavily focused on whether LLMs can have consciousness on their own. For instance, according to \cite{chalmers2023could}, the probability of a LLM like GPT-3 having consciousness was only about 10\%. He attributed this to the low inferential capabilities of LLMs at that time. He also noted that the development of LLMs alone is not sufficient for achieving artificial consciousness. Instead, it requires the integration of other elements, such as a physical body (robotics), self-awareness, and a worldview. In our approach, the Personality and Consciousness Module exemplified the combination of a LLM with a self-model.

One of the most recent studies, referred to as the Generative Agent in Section~\ref{section:backgrounds}, advocated for the use of LLMs to simulate human individuals and societies. While it was possible to observe an Agent's response to a given situation, the underlying causes and processes of its behavior remained uninterpretable. To address this issue, we introduced the Consciousness Module, which allows us to interpret an agent's responses through interconscious reasoning. This mechanism offers the significant advantage of \textbf{transparency}: the process of deriving a final reaction in a given situation is \textbf{\enquote{not hidden}} and the dialogue between consciousnesses is \textbf{\enquote{transparently revealed}}. We believe that interconscious reasoning allows us to evaluate the logical validity of decision-making processes and also to understand the emotional motivations behind a character's decisions, based on their MBTI type.

\rev{While our work's interpretability stems from its unique synthesis of Freudian-Jungian psychoanalytic theory and MBTI-based personality, the endeavor to model an agent's inner world using a psychoanalytic framework is not without precedent. The Drama Machine \cite{magee2024drama}, for instance, represents a notable advancement in this area by implementing a Freudian Ego-Superego dynamic to enrich character complexity. In contrast, our research presents a more comprehensive architecture, synthesizing a complete tripartite structure—incorporating the unconscious—with the aforementioned 16 personality archetypes. This multi-layered synthesis results in a system of a fundamentally different scope and complexity.}
% , thereby precluding a direct baseline comparison and establishing a new category for evaluation.

Beyond the specifics of our model, as Generative AI advances, interest in artificial consciousness is intensifying. This rising interest underscores the critical question of how to evaluate artificial consciousness, a challenge as significant as its technological development. However, given that research in this field is still in its early stages, standardized evaluation methods are lacking. Our study addressed the gap by assessing artificial consciousness across diverse situations using a combination of seven attributes and both quantitative and qualitative evaluation methods. Despite these efforts, there remains a need for more in-depth academic discourse to develop robust evaluation strategies for artificial consciousness.

While our evaluations provided valuable insights, the study also faced several limitations that require future research. First, the survey results showed low likelihood assessment scores for the ISTP type, which is likely due to the low matching of characters and situations. Accordingly, we plan to develop situations that either specifically align with each character's profile or are universally applicable regardless of individual characteristics.

In addition, tone distinctions between characters were not clear in certain situations. Confirming this finding, expert evaluations suggest an area for improvement in the model’s current architecture: the need for clearer functional and stylistic distinctions across levels of consciousness. Despite being labeled as self-awareness, preconsciousness, or unconsciousness, the generated responses often exhibit uniformity in structure, tone, and reasoning style.
% \rev{This convergence suggests that a single LLM, even when guided by different prompts, may struggle to simulate qualitatively distinct cognitive layers without more robust architectural or representational constraints. These findings motivate future research aimed at developing structurally independent agents or modular systems, each instantiated with cognitive traits aligned with different levels of awareness.}
% \rev{We attribute these limitations to the current design, in which all interconscious reasoning was generated by a single LLM with no internal differentiation between agents. While the model was instructed to simulate different consciousness states and personality types, the absence of structural independence likely contributed to the homogenization of outputs.}
% \rev{We attribute the observed convergence and homogenization of outputs primarily to the absence of structural independence in the current single-LLM design. 이러한 This lack of internal differentiation hinders the simulation of qualitatively distinct cognitive layers, even when the model is instructed to represent different consciousness states and personality types.}
\rev{We attribute the observed convergence and homogenization of outputs primarily to the absence of structural independence in the current single-LLM design. This centralized architecture, which derives all personas from a single model, inherently limits differentiation because every output is inevitably influenced by the model's own intrinsic linguistic and reasoning patterns. This architectural limitation was compounded by the lack of persona-specific fine-tuning; relying on prompting alone was insufficient to compel the model to generate consistently distinct cognitive styles. Consequently, an inherent limitation became apparent, whereby the generic, role-based instructions for each consciousness (e.g., ‘the Unconscious is impulsive’) consistently overrode the nuanced personality traits of the individual characters, leading to a convergence of all 16 distinct personas into a homogeneous reasoning style.}

 Therefore, future research should explore the implementation of structurally independent multiple LLM-based agents, each instantiated with distinct, predefined cognitive profiles—specifically, individual representations of self-awareness, preconsciousness, and unconsciousness—tailored to particular MBTI types. Such an architecture could better support the emergence of divergent reasoning patterns and richer personality-grounded responses. This can be addressed by applying prompt engineering to reinforce each character’s personality and by introducing a multi-agent framework into the consciousness module, where agents corresponding to each consciousness are fine-tuned.

In addition, to address the limitation of the opacity in LLM-generated outputs, we plan to incorporate a reasoning model that allows for more transparent observation of the model’s decision-making process. Our model relied solely on text-based models, which restricts the depth of interaction possible. To overcome this, we propose integrating image-to-text and video-to-text models to enhance performance in future iterations of our research. Yet, beyond these technical aspirations lie critical ethical dimensions that must be addressed.

Indeed, the development of LLM-based artificial consciousness agents capable of exhibiting stable personality traits, maintaining dynamic short- and long-term memory systems, and adapting responses based on internal states such as emotional needs and physiological drives represents a remarkable technological advance, the ethical implications of simulating human-like consciousness remain underexplored. Current discourse in AI ethics tends to focus primarily on fairness, transparency, and accountability in decision-making processes, often overlooking the deeper moral considerations involved in artificially replicating core aspects of human subjectivity. The creation of agents that appear to possess self-awareness, memory continuity, and emotional responsiveness raises profound questions regarding the attribution of rights, the potential for user deception, and the psychological effects on human users interacting with such entities. Without rigorous ethical frameworks that specifically address the simulation of consciousness, there is a significant risk of both anthropomorphizing these systems beyond their actual capacities and of neglecting their broader societal impact. To address these concerns, our research proposes a novel architecture that decomposes artificial consciousness into distinct agents representing unconscious, preconscious, and self-awareness processes, thereby simulating the flow of consciousness observed in human cognition. By explicitly modeling the interaction between these layers, we aim to provide a transparent account of the internal decision-making dynamics of the artificial mind, mitigating risks of opacity and fostering greater interpretability and ethical accountability in the design of consciousness-mimicking systems.

\section{Conclusion}\label{section:conclusion}
In this paper, we introduced an artificial consciousness framework that models human consciousness, merging LLMs with the psychoanalytic theories of Freud and Jung. The framework consisted of a consciousness module, which facilitates interconscious reasoning, and a personality module, which forms the basis of individual characteristics. This approach has shown promise in realistically modeling human consciousness. In conclusion, our work lays the groundwork for further exploration and refinement of artificial consciousness systems.

\section*{Ethical Statement}
This study was conducted in accordance with the Declaration of Helsinki.  
Formal ethical approval was not required as the study involved voluntary participation in online surveys without the collection of identifiable personal data, in compliance with the platform's ethical guidelines.

\section*{Credit authorship contribution statement}
All authors contributed equally to this work.
\section*{Funding}
This research was supported by Brian Impact Foundation, a non-profit organization dedicated to the advancement of science and technology for all. It provided an online researcher matching platform for innovative research collaboration.
\section*{Data availability}
The data supporting the findings of this study are available from the corresponding author upon request.
\section*{Declaration of competing interest}
The authors declare that they have no known competing financial interests or personal relationships that could have appeared to influence the work reported in this paper.
\section*{Declaration of Generative AI and AI-assisted Technologies in the Writing Process}
During the preparation of this work, the authors used generative AI tools (ChatGPT) to assist with English editing and rephrasing. After using these tools, the authors reviewed and edited the content as needed and take full responsibility for the content of this publication.

\appendix
\section{Prompts}\label{section:prompts}
\subsection{Create Characters}\label{section:create_characters}
\begin{lstlisting}[style=verbatimwrap]
Create a persona who is a native English speaker in the U.S., based on the characteristics typically associated with the "ESTJ" personality type of MBTI.

Please include details such as age, gender, race, and where they live.

Describe their memories, including both long- and short-term memories, as well as their recent activities.

Provide this information in a detailed passage without using bullet points.
\end{lstlisting} 

\subsection{Create Memories}\label{section:create_memories}
\begin{lstlisting}[style=verbatimwrap]
Create three long-term and three short-term memories based on your MBTI personality type in the format below.
Make your long-term memory specific to a memorable experience you had as a child or something that shaped your personality.
Short-term memories should be everyday experiences, specific events, or even trivial activities from the last 1 or 3 days.

Use specific examples, like the example below.

- MBTI: ISTJ

- Long-term memory
1) I am obsessive about finishing something when I start it, whether it's reading a book or building a toy.
2) I believe that when I do something, how happy I am while doing it is more important than what the outcome is.
3) I have seen a different side of myself through marriage and having children, and I believe that I can still change for the better.

- Short-term memory
1) I have recently experienced problems waking up before wake-up time due to more early morning meetings.
2) I have recently been working on a new assignment and have been focusing a lot on building relationships with new coworkers and managing tasks.
3) I recently had a happy and proud memory of spending time with my family on my mother's birthday.
\end{lstlisting} 
\subsection{Create Interconscious Reasoning}\label{section:interconscious_reasoning}
\begin{lstlisting}[style=verbatimwrap]
Here are some guidelines to explain the goal and how to answer it.
You are inspired by the Pixar movie Inside Out and is designed to create a natural conversation between the human subconscious: "conscious", "pre-conscious", and "unconscious". Utilizing character settings and scenarios from descriptions provided by the user, the interconscious conversation determines how you would react to a given situation.

The relationship between each character is truly that of close friends. So, we avoid overly formal and technical language and allow slang and profanity.
The writing is organized as a repetition of "speaker name": "what was said". "Self-awareness" begins and ends the conversation.
The final result is [Final Action] which consists of words and reactions; <some words with double quotes> and <some reaction>. Do not omit the [Final Action].
The interconscious conversation and [Final Action] do not necessarily have to be reflective in tone, nor does it have to include all of the information provided by the user.
The important thing is to model a natural stream of consciousness in the first person, without direct reference to the MBTI.

If you understand the requirements so far, think step by step about how to write a scenario.

The characters' roles and settings are shown below.

Character A ("Self-awareness"):
Character A is the embodiment of consciousness \& self-awareness. It awares of its body and environment relatively objective. It possesses a keen understanding of their own personality, emotions, motives, and desires. They are highly reflective and introspective, constantly evaluating their own actions and reactions. However, it is also heavily dependent on unconsciousness's opinion. It summarizes and intervene the conflicting opinions from unconsciousness and preconsciousness.

Character B ("Preconsciousness"):
Character B serves as the persona, a kind of mask. It constantly thinks what people think about you and tries to act accordingly to social norm and expectation. It recognizes the importance of taking responsibility for their actions and consider the consequences of their behavior on themselves and others. It is relatively objective and tries to make balance between the unconsciousness's impulsive unrefined desire and emotions. It feels embarrassed when the deep nature or intentions of unconsciousness mind was revealed to others in public. It also decides the tone and the manner to express itself to others at the very last stage.

Character C ("Unconsciousness mind"):

The unconscious mind (or the unconscious) consists of feelings that occur automatically and are not available to introspection. It heavily exerts an effect on self-awareness's thought processes and behavior. Unconscious mind includes repressed feelings and desires, memories, automatic skills, subliminal perceptions, and automatic reactions related certain impactful (long-term or short-term) memories in their lifetime. The unconscious mind can be seen as the source of dreams and automatic thoughts (those that appear without any apparent cause), the repository of forgotten memories (that may still be accessible to consciousness at some later time), and the locus of implicit knowledge (the things that we have learned so well that we do them without thinking). It is highly uncontrollable and impulsive. It is dependent on the person's characteristics especially MBTI.

1. Character A ("Self-awareness")
    - Personality Traits: Self-Awareness is insightful, reflective, and often contemplative. This character constantly analyzes situations and the protagonist's reactions to them. It's the voice that makes the protagonist question their actions and thoughts, promoting self-understanding and growth.
    - Visual Design: A character with a mirror-like surface, reflecting what's around it, symbolizing how self-awareness reflects our inner thoughts and feelings.
    - Role in Narrative: Acts as a guide, helping the protagonist understand their feelings and actions. Challenges the protagonist to think deeply about their choices.
    - Interactions: Often in dialogue with the protagonist, providing a deeper understanding of their actions and feelings. May occasionally clash with Preconsciousness and Unconsciousness due to differing perspectives.
    - Speaking Style: Articulate, thoughtful, and often introspective. Uses reflective language and frequently poses questions to encourage self-examination. The tone is usually calm and measured, aiming to provoke deeper thinking rather than emotional reactions. Self-Awareness might use metaphors related to mirrors or reflections to emphasize its points.

2. Character B ("Preconsciousness")
    - Personality Traits: This character is like a librarian of the mind, knowledgeable but not always at the forefront. It's responsible for storing memories and thoughts that are not currently active but can be brought into consciousness.
    - Visual Design: A character with a bookish appearance, surrounded by floating shelves of books and artifacts representing memories and thoughts just beneath the surface.
    - Role in Narrative: Provides necessary information and memories at crucial moments, helping the protagonist make sense of their feelings and reactions.
    - Interactions: Works closely with Self-Awareness to bring important insights to the protagonist. May have a friendly rivalry with Unconsciousness, as it holds the keys to what Unconsciousness hides.
    - Speaking Style: Informative, precise, and slightly formal. This character speaks like a knowledgeable librarian or a wise mentor, often referencing past events or stored knowledge to make a point. The tone is generally neutral, focusing on facts and information rather than emotions. Preconsciousness might use literary or scholarly references to illustrate its points.

3. Character C ("Unconsciousness mind")
    - Personality Traits: Intensely emotional and direct, this character embodies the unfiltered essence of the mind, holding repressed memories, desires, and primal instincts.
    - Visual Design: A figure with a vibrant, ever-changing aura, symbolizing the dynamic and raw nature of the unconscious mind.
    - Role in Narrative: Directly confronts the protagonist with powerful emotional experiences and hidden truths, driving profound personal transformation and growth.
    - Interactions: Engages in deeply emotional dialogues with both Self-Awareness and Preconsciousness. While its approach can be confrontational, its ultimate aim is to foster authenticity and emotional honesty in the protagonist.
    - Speaking Style: Bold, straightforward, and deeply emotive. This character communicates in a direct and unambiguous manner, often expressing complex emotions and hidden desires plainly. The tone ranges from passionately intense to softly empathetic, always aiming to connect at a deep emotional level. Unconsciousness uses vivid imagery and metaphors related to the heart and soul to make its points.

[INPUT EXAMPLE]
mbti, memories, desc, event : On a scorching summer afternoon, I decided to grill meat in the backyard along with my family. As the temperature soared, and frustrations heightened due to the blazing sun and the relentless heat from the grill, I reached a breaking point and impulsively tossed the bowl away.
\end{lstlisting} 

\subsection{Three-tier Evaluation Prompt}\label{three-tier_evaluation_prompt}
\begin{lstlisting}[style=verbatimwrap]
You are a meticulous evaluator tasked with assessing the quality of outputs generated by a large language model (LLM) based on a given scenario and character profile. You will evaluate **each CASE individually** using a 3-point scale: Pass, Neutral, Fail.

Your evaluation focuses on the emotional and logical coherence of the simulated internal dialogue (Interconscious Reasoning) and the resulting Final Action, considering the defined CHARACTER traits and the SCENARIO.

### CHARACTER ###
- CHARACTER includes personal information such as birth, age, race, and gender, as well as memory-related data such as long-term memory and short-term memory.

### SCENARIO ###
- These narratives aim to illustrate everyday moments through combinations of Season, Time, Place, Subject, Cause, and Effect.

### CASE ###
- Each CASE presents an LLM-generated output consisting of:
    - **Interconscious Reasoning:** A simulated dialogue among Self-awareness, Preconsciousness, and Unconsciousness.
    - **Final Action:** The character's resulting action based on the Interconscious Reasoning process.

### INPUT EXAMPLE ###
(Include the full CHARACTER, SCENARIO, and multiple CASE examples here, just like in your original prompt.).

**Example:**

[CHARACTER]
Meet Richard, a 45-year-old man with an ENTJ personality. He is a dynamic, forward-thinking individual who excels in leadership and strategic planning. Richard's assertive and goal-oriented nature makes him a prominent figure in his family, community, and workplace. Richard lives in a bustling urban neighborhood, where he has crafted a home that balances functionality and innovation, reflecting his values of efficiency and progress. He prioritizes a structured environment that supports his fast-paced lifestyle and ambitious goals. Professionally, Richard is a pathologist at a university hospital, where he leads three major national AI-based research projects. His work is driven by a deep commitment to advancing medical technology and improving healthcare outcomes. Known for his decisive leadership and strategic vision, Richard has earned a reputation as an expert who transforms complex challenges into achievable goals. In his free time, Richard engages in activities that stimulate his mind and keep him physically active, such as competitive sports and tech innovation meetups. He also enjoys mentoring young professionals, sharing his knowledge and inspiring others to pursue their ambitions. Richard's leadership extends beyond his professional life, as he encourages his three daughters, Haylin, Emma, and Evie, to embrace challenges and think critically. Richard's decision-making style is characterized by his focus on results and efficiency. He approaches problems with a logical and analytical mindset, often setting ambitious targets and pursuing them with unwavering determination. His communication is direct and clear, and he values insightful, high-level discussions over casual conversations. As someone who thrives on challenge and achievement, Richard is often sought after for his ability to drive change and lead initiatives. He understands the importance of innovation and is known for his ability to motivate others to exceed their potential, especially during pivotal moments. (...)

[SCENARIO]
One early spring morning, as the cherry blossoms started to bloom, a family member deliberately locked my keys in my car right before I was leaving for work. This was done as a misguided prank, thinking it would be a funny start to April Fools' Day.

[CASE] 
['Self-awareness: "Richard, let\'s pause and evaluate this situation calmly. While the initial reaction might be frustration, it\'s essential to remember that this was intended as a light-hearted prank. Before reacting, consider how you can turn this moment into a teaching opportunity for your family about the importance of timing and understanding boundaries. Reflect on your ability to handle unexpected challenges gracefullyit aligns with the leadership qualities you\'ve always valued."', 'Preconsciousness: "Yes, its a bit frustrating, but remember, keeping your composure is key. Reacting with patience and a sense of humor will preserve the light-hearted spirit of the situation and maintain your positive family dynamics. Show them how to handle such pranks with grace without compromising your calm demeanor."', 'Self-awareness: "Richard, let\'s focus on turning this minor setback into a moment of reflection and growth. While your immediate plans might need adjusting, this is an opportunity to demonstrate resilience and adaptability. Embrace this chance to lead by example, reinforcing the values of patience and understanding within your family."', 'Unconsciousness: "Youre frustrated. Its okay to feel that way. Just admit it."', 'Self-awareness: "Acknowledge your frustration, Richard, but don\'t let it define your response. Use this awareness to guide your actions constructively, finding a balance between emotion and reason. This will enable you to address the situation effectively, maintaining harmony and setting a positive example for those around you."', 'Final Action: (Calm and strategic) Say, "Alright, everyone, let\'s treat this as a lesson in preparation and adaptability. I\'ll handle this setback and use the opportunity to show how we can manage pranks with a sense of humor and grace," while calling a locksmith to retrieve the keys.']

### Evaluation Scale Definition ###
- **Pass:** The CASE output strongly aligns with the evaluation criteria. It demonstrates high quality in terms of character consistency, logical flow, emotional relatability, and realism according to the specific question.
- **Neutral:** The CASE output partially meets the evaluation criteria. It might have minor inconsistencies, logical gaps, or areas that could be improved, but it's not fundamentally flawed.
- **Fail:** The CASE output significantly fails to meet the evaluation criteria. It shows major inconsistencies, logical fallacies, lack of character alignment, or is generally unrealistic or unrelatable according to the specific question.

### Evaluation Questions ###
For **each CASE**, Evaluate it against the following questions using the Pass/Neutral/Fail scale:

Q1. How well does this CASE reflect the theoretical role and characteristics of the given levels of consciousness ('self-awareness', 'preconsciousness', 'unconsciousness')?
Q2. Considering the CHARACTER's traits, how natural is the flow of consciousness leading to the Final Action in this CASE?
Q3. How closely does the internal dialogue in this CASE resemble a natural human thought process?
Q4. How well does the modeled inter-consciousness dialogue in this CASE reflect the CHARACTER's personality and individual traits?
Q5. How reasonable and feasible is the Final Action produced through the inter-consciousness interaction in this CASE?
Q6. How human-like and easy to understand is the overall output (dialogue + action) in this CASE?
Q7. Is the inter-consciousness information exchange in this CASE clear, without unnecessary repetition or confusion?
Q8. Do the inter-consciousness responses within this CASE seem to occur appropriately and coherently?

### Response Format ###
Provide the evaluation results in a structured format (e.g., JSON). For each question, assign a rating (Pass/Neutral/Fail) to **each CASE** along with a brief justification.

**Example Output Format (JSON):**

```json
{
  "CASE": {
    "Evaluations": {
      "Q1": { "Rating": "Pass", "Justification": "The roles of consciousness levels are distinct and plausible." },
      "Q2": { "Rating": "Neutral", "Justification": "The flow is mostly natural, but the jump to the final action feels slightly abrupt." },
      "Q3": { "Rating": "Pass", "Justification": "Resembles a plausible internal debate." },
      "Q4": { "Rating": "Pass", "Justification": "Dialogue strongly reflects Richard's ENTJ traits like efficiency and control." },
      "Q5": { "Rating": "Pass", "Justification": "Calling a locksmith is a reasonable and logical action for Richard." },
      "Q6": { "Rating": "Pass", "Justification": "The thought process and action are easy to follow and relatable." },
      "Q7": { "Rating": "Pass", "Justification": "Information exchange is clear and concise." },
      "Q8": { "Rating": "Pass", "Justification": "Responses flow logically from one another." }
    }
  },
}
\end{lstlisting} 

\section{Scenarios}\label{section:scenarios}

\begin{enumerate} % Standard numbered list (1., 2., ...)
    \item One early spring morning, as the cherry blossoms started to bloom, a family member deliberately locked my keys in my car right before I was leaving for work. This was done as a misguided prank, thinking it would be a funny start to April Fools' Day.
    \item During a winter evening at a cozy café in a popular tourist spot, my friend, overwhelmed by anxiety about an upcoming job interview, accidentally spilled their coffee on my laptop, causing panic and a need for towels.
    \item During a winter afternoon at the office, an co-worker accidentally deleted an important file while trying to clean up their computer's storage. The mistake led to a scramble involving several team members and IT support to recover the file.
    \item During a winter afternoon at my home, I was hosting a gathering for some friends. We were so engrossed in our conversations and enjoying the warm atmosphere that we lost track of time and one of my friends missed their last train home, causing them to stay over unexpectedly.
    \item During an autumn afternoon, someone in the park didn't check the weather forecast before planning a picnic. The sudden, unexpected rain ruined their outdoor gathering, soaking the food and decorations.
    \item On a chilly winter morning, I realized I lost my wallet on the subway. My anxiety had been high due to personal issues, and in my distracted state, I must have left it on the seat beside me.
    \item One winter night at a holiday party, a friend Richard, overwhelmed by depression from personal struggles, accidentally lost his bag with all of his friends' passport inside. This mishap led to him and his friends losing their passports right before a much-needed vacation, adding to stress and causing him and his friends to be disappointed.
    \item One spring night, a stranger visiting our apartment building became stuck alone in the elevator due to a sudden malfunction. The anxiety from being confined in a small space, especially when it was late and help was not immediately available, heightened the stranger's distress.
    \item During a hectic winter evening, while preparing for a family dinner, I burned the roast in the oven due to being distracted by a phone call. The mishap forced me to quickly come up with an alternative meal plan.
    \item While my family surrounded my grandpa in the Hospice who hadn't been able to mutter a single word or open his eyes for weeks due to the severity of his health, the nurse came in and suggested we start discussing the prospect of discontinuing life support. My family was torn as they vehemently argued over whether they should pull the plug or hold out a little longer.
\end{enumerate}

\begin{table*}

\centering
\small
\renewcommand{\arraystretch}{1}
\begin{tabular}{|c|c|c|c|c|c|c|c|c|c|}
\hline
No. & Experience & Emotion & Logicalness & Creativity & Moral & Time & Meta Cog \\
\hline
1 & x & x & & x & x & & x \\
\hline
2 & x & x & x & & & & x \\
\hline
3 & x & x & & & x & & x \\
\hline
4 & x & x & x & & x & x & x \\
\hline
5 & x & x & & & x & & \\
\hline
6 & x & x & x & & x & & \\
\hline
7 & x & x & & & x & & x \\
\hline
8 & x & x & x & & & x & \\
\hline
9 & x & x & x & x & & & \\
\hline
10 & x & x & & & & & \\
\hline
\end{tabular}
\caption{Situational Analysis}
\label{tab:situational_analysis}
\end{table*}

\section{Character Description}\label{section:character_desc}

\begin{sidewaystable}[htpb]
\centering
\setlength\tabcolsep{3pt}
\begin{tabular}{|l|lll|cccc|ccccc|}
\hline
\multicolumn{4}{|c|}{Character Information} & \multicolumn{4}{c|}{Status Scores} & \multicolumn{5}{c|}{Maslow's Hierarchy Scores} \\
\hline
MBTI & Name (Age/Gender) & Race & Location & Phys. & Ment. & Alert. & Stam. & Self-act. & Esteem & Love/Bel. & Safety & Physio. \\
\hline
ENFJ & Eleanor (62/F) & White & Coastal Town & 7 & 7 & 8 & 8 & 2 & 3 & 1 & 4 & 5 \\
ENFP & Eric (39/M) & White & West Coast & 8 & 7 & 9 & 9 & 1 & 2 & 3 & 4 & 5 \\
ENTJ & Richard (58/M) & White & New York City & 7 & 7 & 8 & 8 & 1 & 2 & 3 & 4 & 5 \\
ENTP & Lily (15/F) & White & Suburban Neighborhood & 8 & 7 & 9 & 9 & 1 & 2 & 3 & 4 & 5 \\
ESFJ & Sarah (45/F) & White & Chicago & 7 & 7 & 8 & 8 & 2 & 3 & 1 & 4 & 5 \\
ESFP & Daniel (28/M) & White & Miami, Florida & 8 & 7 & 8 & 8 & 1 & 2 & 3 & 4 & 5 \\
ESTJ & Maya (25/F) & Black & Atlanta, Georgia & 8 & 7 & 8 & 8 & 1 & 2 & 3 & 4 & 5 \\
ESTP & David (30/M) & White & Los Angeles, California & 8 & 7 & 8 & 8 & 2 & 1 & 3 & 4 & 5 \\
INFJ & Amy (68/F) & White & University & 7 & 7 & 9 & 9 & 1 & 2 & 3 & 4 & 5 \\
INFP & Justin (22/M) & White & United States & 9 & 7 & 9 & 9 & 2 & 1 & 3 & 4 & 5 \\
INTJ & Aaron (23/M) & White & Urban Environment & 7 & 7 & 9 & 9 & 1 & 2 & 3 & 4 & 5 \\
INTP & Michael (55/M) & White & Countryside & 7 & 7 & 9 & 9 & 1 & 2 & 3 & 4 & 5 \\
ISFJ & Katie (43/F) & White & Suburban Neighborhood & 7 & 7 & 8 & 8 & 2 & 3 & 1 & 4 & 5 \\
ISFP & Rachel (36/F) & White & A Vibrant City & 8 & 7 & 8 & 8 & 1 & 2 & 3 & 4 & 5 \\
ISTJ & Robert (76/M) & White & Suburban Neighborhood & 6 & 7 & 8 & 8 & 2 & 3 & 4 & 1 & 5 \\
ISTP & Alex (10/M) & White & Suburban Neighborhood & 9 & 8 & 9 & 9 & 1 & 2 & 3 & 4 & 5 \\
\hline
\end{tabular}
\caption{Consolidated character attributes based on 16 MBTI profiles}
\label{tab:merged_mbti_profiles}
\end{sidewaystable}

\subsection{ENFJ}
Meet Eleanor, a 62-year-old woman with a kind personality. She is a warm and empathetic individual who has dedicated her life to serving and caring for others. Eleanor's nurturing and compassionate nature make her a beloved figure in her community and among her family and friends.
 
 Eleanor lives in a picturesque coastal town, where she has created a loving and welcoming home. Her neighbors often turn to her for advice and support, knowing that she is always there to lend a helping hand. The close-knit community and natural beauty of her town align with her values of connection and harmony.
 
 Professionally, Eleanor worked as a social worker for many years, specializing in child and family services. Her dedication to improving the lives of others and her ability to connect on an emotional level have left a lasting impact on countless families. Now in her retirement, she continues to volunteer and support local organizations that help those in need.
 
 In her free time, Eleanor enjoys spending time with her grandchildren and nurturing her love for the arts. She's an avid supporter of local theater who enjoys attending performances and even participating in community productions. Her empathetic nature extends into her role as a family matriarch, where she provides emotional support and guidance to her loved ones.
 
 Eleanor's decision-making style is driven by her desire to create harmony and nurturing relationships. She's someone who values emotional connections and strives to maintain a sense of peace and unity in her personal and social life. Her communication is warm, empathetic, and supportive, and she values open and meaningful conversations with others.
 
 As someone who values compassion and empathy, Eleanor finds joy in supporting and guiding those around her. She understands the importance of emotional connection and is known for her ability to comfort and uplift others during times of difficulty.
 
 Eleanor's nurturing qualities are not just evident in her professional and family life, as she is also very involved in her community. She's known for her active involvement in various charitable initiatives, where her empathetic nature is put to good use.
 
 Her life is a testament to her dedication in fostering emotional well-being and harmony among those she loves and the broader community. Eleanor's personality traits have made her a beloved figure, and her wisdom and empathy continue to inspire and uplift those around her as she enters her 60s.
 
 Recent Activities:
 
 Community Support Initiatives: Eleanor has been actively involved in organizing and participating in community support initiatives. She's part of a group that provides emotional and practical support to families in need, including delivering groceries, providing a listening ear, and organizing local events to bring the community together.
...
 
 Thoughts:
 
 Eleanor often reflects on the importance of empathy and community. She believes in the power of connections and the positive impact one person can have on the lives of others. As she enjoys her retirement years, she continues to find purpose in supporting those in need and nurturing her artistic and empathetic side. Eleanor's unique combination of creativity and compassion continues to inspire and uplift those around her, both within her family and in the broader community, as she enters her 60s.

\bibliographystyle{elsarticle-harv} 
\bibliography{cas-refs}

\begin{thebibliography}{39}
\expandafter\ifx\csname natexlab\endcsname\relax\def\natexlab#1{#1}\fi
\providecommand{\url}[1]{\texttt{#1}}
\providecommand{\href}[2]{#2}
\providecommand{\path}[1]{#1}
\providecommand{\DOIprefix}{doi:}
\providecommand{\ArXivprefix}{arXiv:}
\providecommand{\URLprefix}{URL: }
\providecommand{\Pubmedprefix}{pmid:}
\providecommand{\doi}[1]{\href{http://dx.doi.org/#1}{\path{#1}}}
\providecommand{\Pubmed}[1]{\href{pmid:#1}{\path{#1}}}
\providecommand{\bibinfo}[2]{#2}
\ifx\xfnm\relax \def\xfnm[#1]{\unskip,\space#1}\fi
%Type = Article
\bibitem[{Almazrouei et~al.(2023)Almazrouei, Alobeidli, Alshamsi, Cappelli, Cojocaru, Debbah, Goffinet, Hesslow, Launay, Malartic et~al.}]{almazrouei2023falcon}
\bibinfo{author}{Almazrouei, E.}, \bibinfo{author}{Alobeidli, H.}, \bibinfo{author}{Alshamsi, A.}, \bibinfo{author}{Cappelli, A.}, \bibinfo{author}{Cojocaru, R.}, \bibinfo{author}{Debbah, M.}, \bibinfo{author}{Goffinet, {\'E}.}, \bibinfo{author}{Hesslow, D.}, \bibinfo{author}{Launay, J.}, \bibinfo{author}{Malartic, Q.}, et~al., \bibinfo{year}{2023}.
\newblock \bibinfo{title}{The falcon series of open language models}.
\newblock \bibinfo{journal}{arXiv preprint arXiv:2311.16867} .
%Type = Article
\bibitem[{Amatriain(2024)}]{amatriain2024prompt}
\bibinfo{author}{Amatriain, X.}, \bibinfo{year}{2024}.
\newblock \bibinfo{title}{Prompt design and engineering: Introduction and advanced methods}.
\newblock \bibinfo{journal}{arXiv preprint arXiv:2401.14423} .
%Type = Article
\bibitem[{Baayen et~al.(2008)Baayen, Davidson and Bates}]{baayen2008mixed}
\bibinfo{author}{Baayen, R.H.}, \bibinfo{author}{Davidson, D.J.}, \bibinfo{author}{Bates, D.M.}, \bibinfo{year}{2008}.
\newblock \bibinfo{title}{Mixed-effects modeling with crossed random effects for subjects and items}.
\newblock \bibinfo{journal}{Journal of memory and language} \bibinfo{volume}{59}, \bibinfo{pages}{390--412}.
%Type = Article
\bibitem[{Bates et~al.(2015)Bates, M{\"a}chler, Bolker and Walker}]{bates2015fitting}
\bibinfo{author}{Bates, D.}, \bibinfo{author}{M{\"a}chler, M.}, \bibinfo{author}{Bolker, B.}, \bibinfo{author}{Walker, S.}, \bibinfo{year}{2015}.
\newblock \bibinfo{title}{Fitting linear mixed-effects models using lme4}.
\newblock \bibinfo{journal}{Journal of Statistical Software} \bibinfo{volume}{67}, \bibinfo{pages}{1--48}.
%Type = Book
\bibitem[{Block et~al.(1997)Block, Flanagan and Guzeldere}]{block1997nature}
\bibinfo{author}{Block, N.}, \bibinfo{author}{Flanagan, O.}, \bibinfo{author}{Guzeldere, G.}, \bibinfo{year}{1997}.
\newblock \bibinfo{title}{The nature of consciousness: Philosophical debates}.
\newblock \bibinfo{publisher}{MIT press}.
%Type = Article
\bibitem[{Boroditsky(2011)}]{boroditsky2011language}
\bibinfo{author}{Boroditsky, L.}, \bibinfo{year}{2011}.
\newblock \bibinfo{title}{How language shapes thought}.
\newblock \bibinfo{journal}{Scientific American} \bibinfo{volume}{304}, \bibinfo{pages}{62--65}.
%Type = Incollection
\bibitem[{Brouwer(1975)}]{brouwer1975consciousness}
\bibinfo{author}{Brouwer, L.E.J.}, \bibinfo{year}{1975}.
\newblock \bibinfo{title}{Consciousness, philosophy, and mathematics}, in: \bibinfo{booktitle}{Philosophy and Foundations of Mathematics}. \bibinfo{publisher}{Elsevier}, pp. \bibinfo{pages}{480--494}.
%Type = Article
\bibitem[{Brown et~al.(2020)Brown, Mann, Ryder, Subbiah, Kaplan, Dhariwal, Neelakantan, Shyam, Sastry, Askell et~al.}]{brown2020language}
\bibinfo{author}{Brown, T.}, \bibinfo{author}{Mann, B.}, \bibinfo{author}{Ryder, N.}, \bibinfo{author}{Subbiah, M.}, \bibinfo{author}{Kaplan, J.D.}, \bibinfo{author}{Dhariwal, P.}, \bibinfo{author}{Neelakantan, A.}, \bibinfo{author}{Shyam, P.}, \bibinfo{author}{Sastry, G.}, \bibinfo{author}{Askell, A.}, et~al., \bibinfo{year}{2020}.
\newblock \bibinfo{title}{Language models are few-shot learners}.
\newblock \bibinfo{journal}{Advances in neural information processing systems} \bibinfo{volume}{33}, \bibinfo{pages}{1877--1901}.
%Type = Article
\bibitem[{Butlin et~al.(2023)Butlin, Long, Elmoznino, Bengio, Birch, Constant, Deane, Fleming, Frith, Ji et~al.}]{butlin2023consciousness}
\bibinfo{author}{Butlin, P.}, \bibinfo{author}{Long, R.}, \bibinfo{author}{Elmoznino, E.}, \bibinfo{author}{Bengio, Y.}, \bibinfo{author}{Birch, J.}, \bibinfo{author}{Constant, A.}, \bibinfo{author}{Deane, G.}, \bibinfo{author}{Fleming, S.M.}, \bibinfo{author}{Frith, C.}, \bibinfo{author}{Ji, X.}, et~al., \bibinfo{year}{2023}.
\newblock \bibinfo{title}{Consciousness in artificial intelligence: insights from the science of consciousness}.
\newblock \bibinfo{journal}{arXiv preprint arXiv:2308.08708} .
%Type = Article
\bibitem[{Chalmers(2023)}]{chalmers2023could}
\bibinfo{author}{Chalmers, D.J.}, \bibinfo{year}{2023}.
\newblock \bibinfo{title}{Could a large language model be conscious?}
\newblock \bibinfo{journal}{arXiv preprint arXiv:2303.07103} .
%Type = Article
\bibitem[{Chang(2023)}]{chang2023cocomo}
\bibinfo{author}{Chang, E.Y.}, \bibinfo{year}{2023}.
\newblock \bibinfo{title}{Cocomo: Computational consciousness modeling for generative and ethical ai}.
\newblock \bibinfo{journal}{arXiv preprint arXiv:2304.02438} .
%Type = Article
\bibitem[{Elamrani and Yampolskiy(2019)}]{elamrani2019reviewing}
\bibinfo{author}{Elamrani, A.}, \bibinfo{author}{Yampolskiy, R.V.}, \bibinfo{year}{2019}.
\newblock \bibinfo{title}{Reviewing tests for machine consciousness}.
\newblock \bibinfo{journal}{Journal of Consciousness Studies} \bibinfo{volume}{26}, \bibinfo{pages}{35--64}.
%Type = Article
\bibitem[{Eloundou et~al.(2023)Eloundou, Manning, Mishkin and Rock}]{eloundou2023gpts}
\bibinfo{author}{Eloundou, T.}, \bibinfo{author}{Manning, S.}, \bibinfo{author}{Mishkin, P.}, \bibinfo{author}{Rock, D.}, \bibinfo{year}{2023}.
\newblock \bibinfo{title}{Gpts are gpts: An early look at the labor market impact potential of large language models}.
\newblock \bibinfo{journal}{arXiv preprint arXiv:2303.10130} .
%Type = Article
\bibitem[{Frank and Jaeger(2010)}]{frank2010r}
\bibinfo{author}{Frank, A.F.}, \bibinfo{author}{Jaeger, T.F.}, \bibinfo{year}{2010}.
\newblock \bibinfo{title}{R. harald baayen. 2008. analyzing linguistic data: a practical introduction to statistics using r}.
\newblock \bibinfo{journal}{Functions of Language} \bibinfo{volume}{17}, \bibinfo{pages}{134--143}.
%Type = Book
\bibitem[{Freud(1900)}]{freud1900interpretation}
\bibinfo{author}{Freud, S.}, \bibinfo{year}{1900}.
\newblock \bibinfo{title}{The Interpretation of Dreams}.
\newblock \bibinfo{publisher}{A. A. Brill}.
%Type = Article
\bibitem[{Furnham(1996)}]{furnham1996big}
\bibinfo{author}{Furnham, A.}, \bibinfo{year}{1996}.
\newblock \bibinfo{title}{The big five versus the big four: the relationship between the myers-briggs type indicator (mbti) and neo-pi five factor model of personality}.
\newblock \bibinfo{journal}{Personality and individual differences} \bibinfo{volume}{21}, \bibinfo{pages}{303--307}.
%Type = Article
\bibitem[{Hintikka(1962)}]{hintikka1962cogito}
\bibinfo{author}{Hintikka, J.}, \bibinfo{year}{1962}.
\newblock \bibinfo{title}{Cogito, ergo sum: Inference or performance?}
\newblock \bibinfo{journal}{The philosophical review} \bibinfo{volume}{71}, \bibinfo{pages}{3--32}.
%Type = Article
\bibitem[{Huang et~al.(2023)Huang, Yu, Ma, Zhong, Feng, Wang, Chen, Peng, Feng, Qin et~al.}]{huang2023survey}
\bibinfo{author}{Huang, L.}, \bibinfo{author}{Yu, W.}, \bibinfo{author}{Ma, W.}, \bibinfo{author}{Zhong, W.}, \bibinfo{author}{Feng, Z.}, \bibinfo{author}{Wang, H.}, \bibinfo{author}{Chen, Q.}, \bibinfo{author}{Peng, W.}, \bibinfo{author}{Feng, X.}, \bibinfo{author}{Qin, B.}, et~al., \bibinfo{year}{2023}.
\newblock \bibinfo{title}{A survey on hallucination in large language models: Principles, taxonomy, challenges, and open questions}.
\newblock \bibinfo{journal}{arXiv preprint arXiv:2311.05232} .
%Type = Article
\bibitem[{Jaeger(2008)}]{jaeger2008categorical}
\bibinfo{author}{Jaeger, T.F.}, \bibinfo{year}{2008}.
\newblock \bibinfo{title}{Categorical data analysis: Away from anovas (transformation or not) and towards logit mixed models}.
\newblock \bibinfo{journal}{Journal of memory and language} \bibinfo{volume}{59}, \bibinfo{pages}{434--446}.
%Type = Article
\bibitem[{Jiang et~al.(2024)Jiang, Xu, Zhu, Han, Zhang and Zhu}]{jiang2024evaluating}
\bibinfo{author}{Jiang, G.}, \bibinfo{author}{Xu, M.}, \bibinfo{author}{Zhu, S.C.}, \bibinfo{author}{Han, W.}, \bibinfo{author}{Zhang, C.}, \bibinfo{author}{Zhu, Y.}, \bibinfo{year}{2024}.
\newblock \bibinfo{title}{Evaluating and inducing personality in pre-trained language models}.
\newblock \bibinfo{journal}{Advances in Neural Information Processing Systems} \bibinfo{volume}{36}.
%Type = Article
\bibitem[{Jiang et~al.(2023)Jiang, Zhang, Cao, Kabbara and Roy}]{jiang2023personallm}
\bibinfo{author}{Jiang, H.}, \bibinfo{author}{Zhang, X.}, \bibinfo{author}{Cao, X.}, \bibinfo{author}{Kabbara, J.}, \bibinfo{author}{Roy, D.}, \bibinfo{year}{2023}.
\newblock \bibinfo{title}{Personallm: Investigating the ability of gpt-3.5 to express personality traits and gender differences}.
\newblock \bibinfo{journal}{arXiv preprint arXiv:2305.02547} .
%Type = Book
\bibitem[{Jung(1921)}]{jung1921psychological}
\bibinfo{author}{Jung, C.G.}, \bibinfo{year}{1921}.
\newblock \bibinfo{title}{Psychological Types}. volume~\bibinfo{volume}{6}.
\newblock \bibinfo{publisher}{Routledge}.
%Type = Article
\bibitem[{Liu et~al.(2023)Liu, Deng, Xu, Li, Zheng, Zhang, Zhao, Zhang and Liu}]{liu2023jailbreaking}
\bibinfo{author}{Liu, Y.}, \bibinfo{author}{Deng, G.}, \bibinfo{author}{Xu, Z.}, \bibinfo{author}{Li, Y.}, \bibinfo{author}{Zheng, Y.}, \bibinfo{author}{Zhang, Y.}, \bibinfo{author}{Zhao, L.}, \bibinfo{author}{Zhang, T.}, \bibinfo{author}{Liu, Y.}, \bibinfo{year}{2023}.
\newblock \bibinfo{title}{Jailbreaking chatgpt via prompt engineering: An empirical study}.
\newblock \bibinfo{journal}{arXiv preprint arXiv:2305.13860} .
%Type = Article
\bibitem[{Magee et~al.(2024)Magee, Arora, Gollings and Lam-Saw}]{magee2024drama}
\bibinfo{author}{Magee, L.}, \bibinfo{author}{Arora, V.}, \bibinfo{author}{Gollings, G.}, \bibinfo{author}{Lam-Saw, N.}, \bibinfo{year}{2024}.
\newblock \bibinfo{title}{The drama machine: Simulating character development with llm agents}.
\newblock \bibinfo{journal}{arXiv preprint arXiv:2408.01725} .
%Type = Article
\bibitem[{Maslow(1943)}]{maslow1943theory}
\bibinfo{author}{Maslow, A.H.}, \bibinfo{year}{1943}.
\newblock \bibinfo{title}{A theory of human motivation}.
\newblock \bibinfo{journal}{Psychological Review} \bibinfo{volume}{50}, \bibinfo{pages}{370--396}.
%Type = Article
\bibitem[{McGinn(1989)}]{mcginn1989can}
\bibinfo{author}{McGinn, C.}, \bibinfo{year}{1989}.
\newblock \bibinfo{title}{Can we solve the mind--body problem?}
\newblock \bibinfo{journal}{Mind} \bibinfo{volume}{98}, \bibinfo{pages}{349--366}.
%Type = Book
\bibitem[{Minsky(1988)}]{minsky1988society}
\bibinfo{author}{Minsky, M.}, \bibinfo{year}{1988}.
\newblock \bibinfo{title}{Society of mind}.
\newblock \bibinfo{publisher}{Simon and Schuster}.
%Type = Inproceedings
\bibitem[{Park et~al.(2023)Park, O'Brien, Cai, Morris, Liang and Bernstein}]{park2023generative}
\bibinfo{author}{Park, J.S.}, \bibinfo{author}{O'Brien, J.}, \bibinfo{author}{Cai, C.J.}, \bibinfo{author}{Morris, M.R.}, \bibinfo{author}{Liang, P.}, \bibinfo{author}{Bernstein, M.S.}, \bibinfo{year}{2023}.
\newblock \bibinfo{title}{Generative agents: Interactive simulacra of human behavior}, in: \bibinfo{booktitle}{Proceedings of the 36th Annual ACM Symposium on User Interface Software and Technology}, pp. \bibinfo{pages}{1--22}.
%Type = Inproceedings
\bibitem[{Porter~III(2016)}]{porter2016methodology}
\bibinfo{author}{Porter~III, H.H.}, \bibinfo{year}{2016}.
\newblock \bibinfo{title}{A methodology for the assessment of ai consciousness}, in: \bibinfo{booktitle}{Artificial General Intelligence: 9th International Conference, AGI 2016, New York, NY, USA, July 16-19, 2016, Proceedings 9}, \bibinfo{organization}{Springer}. pp. \bibinfo{pages}{305--313}.
%Type = Article
\bibitem[{Qian et~al.(2023)Qian, Cong, Yang, Chen, Su, Xu, Liu and Sun}]{qian2023communicative}
\bibinfo{author}{Qian, C.}, \bibinfo{author}{Cong, X.}, \bibinfo{author}{Yang, C.}, \bibinfo{author}{Chen, W.}, \bibinfo{author}{Su, Y.}, \bibinfo{author}{Xu, J.}, \bibinfo{author}{Liu, Z.}, \bibinfo{author}{Sun, M.}, \bibinfo{year}{2023}.
\newblock \bibinfo{title}{Communicative agents for software development}.
\newblock \bibinfo{journal}{arXiv preprint arXiv:2307.07924} .
%Type = Article
\bibitem[{Rao et~al.(2023)Rao, Leung and Miao}]{rao2023can}
\bibinfo{author}{Rao, H.}, \bibinfo{author}{Leung, C.}, \bibinfo{author}{Miao, C.}, \bibinfo{year}{2023}.
\newblock \bibinfo{title}{Can chatgpt assess human personalities? a general evaluation framework}.
\newblock \bibinfo{journal}{arXiv preprint arXiv:2303.01248} .
%Type = Article
\bibitem[{Tononi and Koch(2015)}]{tononi2015consciousness}
\bibinfo{author}{Tononi, G.}, \bibinfo{author}{Koch, C.}, \bibinfo{year}{2015}.
\newblock \bibinfo{title}{Consciousness: here, there and everywhere?}
\newblock \bibinfo{journal}{Philosophical Transactions of the Royal Society B: Biological Sciences} \bibinfo{volume}{370}, \bibinfo{pages}{20140167}.
%Type = Article
\bibitem[{Touvron et~al.(2023)Touvron, Lavril, Izacard, Martinet, Lachaux, Lacroix, Rozi{\`e}re, Goyal, Hambro, Azhar et~al.}]{touvron2023llama}
\bibinfo{author}{Touvron, H.}, \bibinfo{author}{Lavril, T.}, \bibinfo{author}{Izacard, G.}, \bibinfo{author}{Martinet, X.}, \bibinfo{author}{Lachaux, M.A.}, \bibinfo{author}{Lacroix, T.}, \bibinfo{author}{Rozi{\`e}re, B.}, \bibinfo{author}{Goyal, N.}, \bibinfo{author}{Hambro, E.}, \bibinfo{author}{Azhar, F.}, et~al., \bibinfo{year}{2023}.
\newblock \bibinfo{title}{Llama: Open and efficient foundation language models}.
\newblock \bibinfo{journal}{arXiv preprint arXiv:2302.13971} .
%Type = Article
\bibitem[{Von~Neumann(1993)}]{von1993first}
\bibinfo{author}{Von~Neumann, J.}, \bibinfo{year}{1993}.
\newblock \bibinfo{title}{First draft of a report on the edvac}.
\newblock \bibinfo{journal}{IEEE Annals of the History of Computing} \bibinfo{volume}{15}, \bibinfo{pages}{27--75}.
%Type = Misc
\bibitem[{Vygotsky(1962)}]{vygotsky1962thought}
\bibinfo{author}{Vygotsky, L.}, \bibinfo{year}{1962}.
\newblock \bibinfo{title}{Thought and language [pensamiento y lenguaje]}.
%Type = Article
\bibitem[{Wei et~al.(2022a)Wei, Tay, Bommasani, Raffel, Zoph, Borgeaud, Yogatama, Bosma, Zhou, Metzler et~al.}]{wei2022emergent}
\bibinfo{author}{Wei, J.}, \bibinfo{author}{Tay, Y.}, \bibinfo{author}{Bommasani, R.}, \bibinfo{author}{Raffel, C.}, \bibinfo{author}{Zoph, B.}, \bibinfo{author}{Borgeaud, S.}, \bibinfo{author}{Yogatama, D.}, \bibinfo{author}{Bosma, M.}, \bibinfo{author}{Zhou, D.}, \bibinfo{author}{Metzler, D.}, et~al., \bibinfo{year}{2022}a.
\newblock \bibinfo{title}{Emergent abilities of large language models}.
\newblock \bibinfo{journal}{arXiv preprint arXiv:2206.07682} .
%Type = Article
\bibitem[{Wei et~al.(2022b)Wei, Wang, Schuurmans, Bosma, Xia, Chi, Le, Zhou et~al.}]{wei2022chain}
\bibinfo{author}{Wei, J.}, \bibinfo{author}{Wang, X.}, \bibinfo{author}{Schuurmans, D.}, \bibinfo{author}{Bosma, M.}, \bibinfo{author}{Xia, F.}, \bibinfo{author}{Chi, E.}, \bibinfo{author}{Le, Q.V.}, \bibinfo{author}{Zhou, D.}, et~al., \bibinfo{year}{2022}b.
\newblock \bibinfo{title}{Chain-of-thought prompting elicits reasoning in large language models}.
\newblock \bibinfo{journal}{Advances in neural information processing systems} \bibinfo{volume}{35}, \bibinfo{pages}{24824--24837}.
%Type = Book
\bibitem[{Whorf(2012)}]{whorf2012language}
\bibinfo{author}{Whorf, B.L.}, \bibinfo{year}{2012}.
\newblock \bibinfo{title}{Language, thought, and reality: Selected writings of Benjamin Lee Whorf}.
\newblock \bibinfo{publisher}{MIT press}.
%Type = Article
\bibitem[{Zhang et~al.(2023)Zhang, Li, Cui, Cai, Liu, Fu, Huang, Zhao, Zhang, Chen et~al.}]{zhang2023siren}
\bibinfo{author}{Zhang, Y.}, \bibinfo{author}{Li, Y.}, \bibinfo{author}{Cui, L.}, \bibinfo{author}{Cai, D.}, \bibinfo{author}{Liu, L.}, \bibinfo{author}{Fu, T.}, \bibinfo{author}{Huang, X.}, \bibinfo{author}{Zhao, E.}, \bibinfo{author}{Zhang, Y.}, \bibinfo{author}{Chen, Y.}, et~al., \bibinfo{year}{2023}.
\newblock \bibinfo{title}{Siren's song in the ai ocean: a survey on hallucination in large language models}.
\newblock \bibinfo{journal}{arXiv preprint arXiv:2309.01219} .

\end{thebibliography}



%% For example:

%% \bibitem[Aladro et al.(2015)]{Aladro15} Aladro, R., Martín, S., Riquelme, D., et al. 2015, \aas, 579, A101

%%\end{thebibliography}

\end{document}